%% file: main.tex
\documentclass[10pt,twocolumn,letterpaper]{article}

\usepackage{cvpr}              


\definecolor{cvprblue}{rgb}{0.21,0.49,0.74}
\usepackage[pagebackref,breaklinks,colorlinks,allcolors=cvprblue]{hyperref}

\def\paperID{*****} 
\def\confName{CVPR}
\def\confYear{2025}

\title{The 1st Solution for MOSEv1 Challenge on LSVOS 2025: CGFSeg}

\author{
Tingmin Li\textsuperscript{1} \quad
Yixuan Li\textsuperscript{1} \quad
Yang Yang\textsuperscript{1} \\
\textsuperscript{1}Nanjing University of Science and Technology \\
}

\begin{document}
\maketitle

\input{sec/0_abstract}

\input{sec/1_intro}
\input{sec/2_method}

\input{sec/3_experiments}
\input{sec/4_conclusion}

{
    \small
    \bibliographystyle{ieeenat_fullname}
    \bibliography{main}
}


\end{document}

%% file: sec/0_abstract.tex
\begin{abstract}
Video Object Segmentation (VOS) aims to track and segment specific objects across entire video sequences, yet it remains highly challenging under complex real-world scenarios. The MOSEv1 and LVOS dataset, adopted in the MOSEv1 challenge on LSVOS 2025, which is specifically designed to enhance the robustness of VOS models in complex real-world scenarios, including long-term object disappearances and reappearances, as well as the presence of small and inconspicuous objects. In this paper, we present our improved method, Confidence-Guided Fusion Segmentation (CGFSeg), for the VOS task in the MOSEv1 Challenge. During training, the feature extractor of SAM2 is frozen, while the remaining components are fine-tuned to preserve strong feature extraction ability and improve segmentation accuracy. In the inference stage, we introduce a pixel-check strategy that progressively refines predictions by exploiting complementary strengths of multiple models, thereby yielding robust final masks. As a result, our method achieves a J\&F score of 86.37\% on the test set, ranking 1st in the MOSEv1 Challenge at LSVOS 2025. These results highlight the effectiveness of our approach in addressing the challenges of VOS task in complex scenarios.
\end{abstract}

%% file: sec/1_intro.tex
\section{Introduction}
The 7th Large-scale Video Object Segmentation (LSVOS) Challenge is designed to advance the state of the art in video object segmentation by encouraging the development of models that can generalize to more diverse and challenging real-world scenarios. The competition includes three tracks: Video Object Segmentation (Classic VOS), Referring Video Object Segmentation (RVOS), and Complex Video Object Segmentation (MOSEv2). The VOS track leverages the LVOS\cite{Hong_2023_ICCV} and MOSEv1\cite{ding2023mose} datasets to study video object segmentation in challenging environments. Specifically, LVOS is designed for long-term video, aiming to simulate long-term object reappearance and visually similar objects in extended temporal span scenarios, while MOSEv1 focuses on intricate scenes involving small and inconspicuous objects, heavy occlusions, and crowded environments. The RVOS track employs the MeViS\cite{MeViS,ding2025mevis} dataset to investigate motion-driven object grounding guided by natural language descriptions. This dataset\cite{ding2025mevis} contains a large number of motion-related expressions, covering 8,171 objects in 2,006 videos of complex scenarios, which are used to specify target objects in complex environments. The MOSEv2 track utilizes the MOSEv2\cite{ding2025mosev2challengingdatasetvideo} dataset, emphasizing segmentation in more complex scenes with frequent object disappearance and reappearance, severe occlusions, smaller targets, as well as new challenges such as adverse weather, low-light conditions, multi-shot sequences, and camouflage. This paper primarily details our method and experimental results for the Classic VOS track.

\begin{figure}[t]  
  \centering
\includegraphics[width=0.48\textwidth]{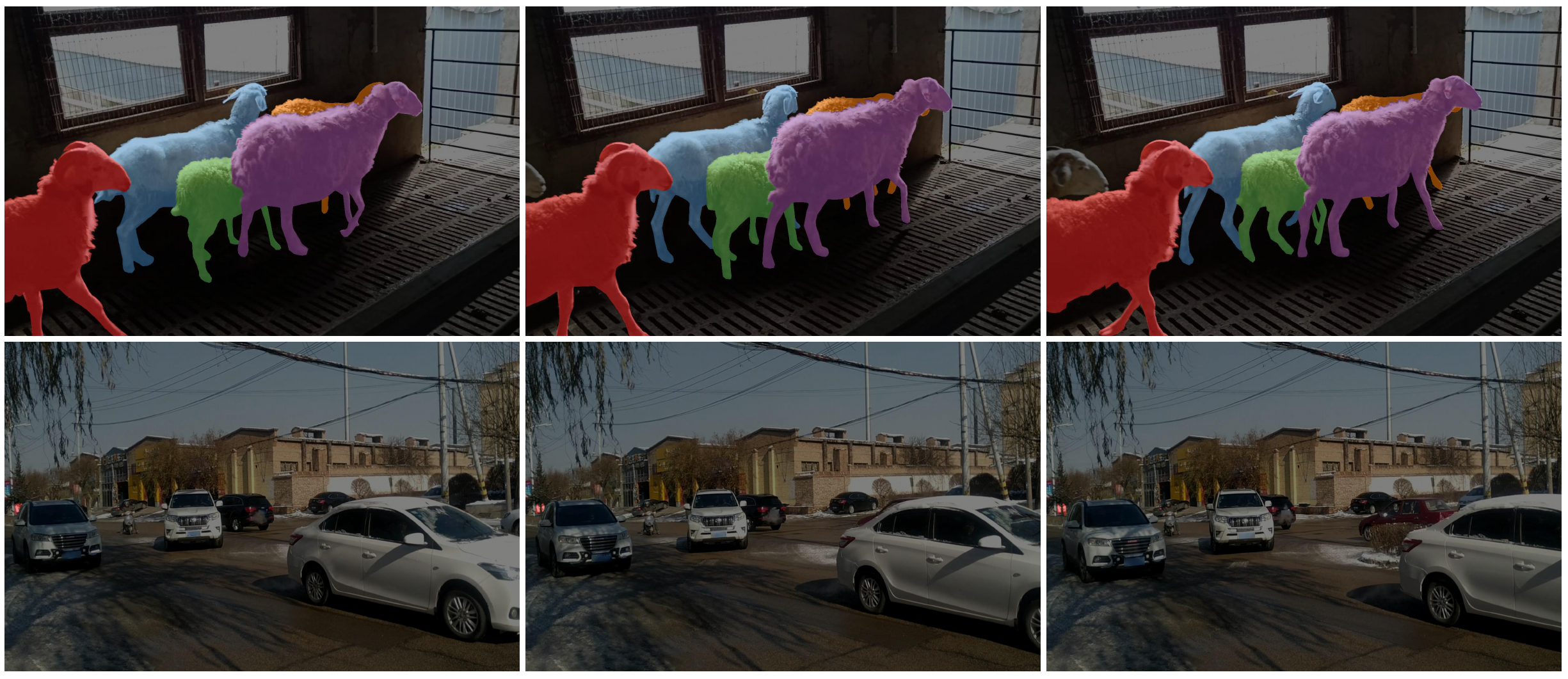}
  \caption{Examples of video clips from the co\textbf{M}plex video \textbf{O}bject \textbf{SE}gmentation (MOSEv1) dataset. The first row illustrates appearance ambiguities caused by occlusion in crowded scenes, while the second row highlights challenges of small object motion.}
  \label{fig:introduction}
\end{figure}

Video Object Segmentation (VOS)\cite{Benchmark:conf/cvpr/PerazziPMGGS16, one_shot:conf/cvpr/CaellesMPLCG17, distractor:conf/cvpr/VidenovicLK25} is a fundamental task in computer vision, aiming to consistently segment specific target objects across entire video sequences. VOS is generally divided into semi-supervised and unsupervised settings\cite{xu2018youtube}. The Classic VOS track in this challenge focuses on the semi-supervised setting, where the objective is to segment target objects throughout a video sequence, given only the ground-truth masks in the first frame. To thoroughly evaluate the performance of the VOS methods in realistic and complex conditions, the competition introduces the MOSEv1 dataset and LVOS dataset, which present challenges including long-term object reappearance, severe occlusions, and intricate motion patterns. Figure \ref{fig:introduction} shows some typical challenges encountered in the MOSEv1 dataset.
\begin{figure*}[t]
\centering \includegraphics[width=0.9\textwidth]{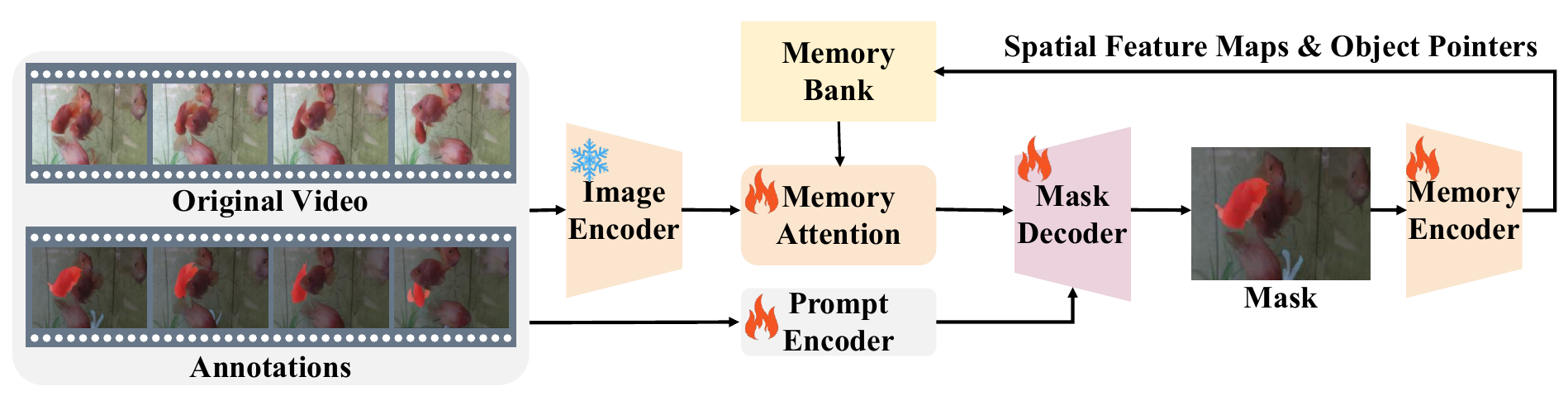} \caption{Framework of our method for the training stage.} \label{fig:training} \end{figure*}
Current mainstream VOS approaches commonly employ memory architectures to store and utilize object information from historical frames\cite{oh2019stm,memort_method_2:conf/nips/ChengTT21,memort_method_3:conf/eccv/SeongHK20}. For instance,  XMem\cite{cheng2022xmem} introduces a hierarchical pixel-level memory mechanism that enables efficient information storage and retrieval, substantially improving the stability of long-video segmentation. Building on this, Cutie\cite{cheng2024putting} further incorporates object-level attention, elevating feature modeling from the pixel to the object level and thereby enhancing the discrimination of complex occlusions and interference from similar objects. SAM2 \cite{ravi2024sam2} incorporates a memory attention mechanism that enables the model to leverage information from historical frames, effectively capturing temporal dependencies, improving the model’s ability to maintain object consistency across frames. SAM2Long\cite{ding2025sam2long} builds upon SAM 2 by proposing a multipath memory tree structure and uncertainty mechanisms to mitigate error accumulation in long videos, which further improves segmentation accuracy. However, despite these approaches achieving strong performance on conventional benchmarks, they suffer significant degradation under more challenging scenarios. For example, the J\&F score of XMem \cite{cheng2022xmem} on the DAVIS17\cite{davis17:journals/corr/Pont-TusetPCASG17} dataset is 86.2\%, but it  drops sharply to 57.6\% on the MOSEv1 dataset. Therefore, enhancing model robustness under such challenging scenarios is critical for improving the performance of VOS task.
\begin{figure*}[t]
\centering \includegraphics[width=0.9\textwidth]{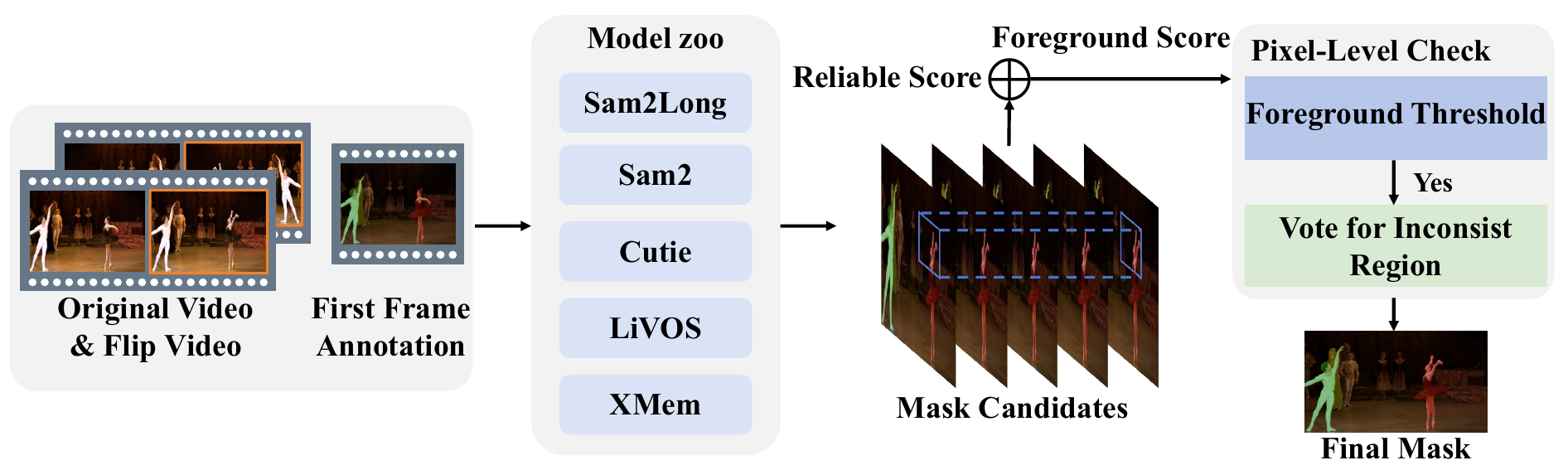} \caption{Framework of our method for the inference stage.} \label{fig:test} \end{figure*}

To address these challenge, we propose a Confidence-Guided Fusion Segmentation (CGFSeg) strategy that leverages the complementary strengths of multiple VOS models to obtain more reliable segmentation results. Specifically, during the training phase, the SAM2 model is fine-tuned on the complex motion patterns provided by MOSEv1 and LVOS, enabling it to better handle occlusions and long-term reappearance problems in challenging real-world  scenarios. During inference, a pixel-level check strategy is employed to fuse the outputs of multiple models based on confidence scores and to determine the foreground for each pixel, ensuring that the most reliable foreground regions are retained. Subsequently, for regions with inconsistent predictions, a voting mechanism is applied to assign object IDs to each pixel, addressing object overlap issues and further improving segmentation accuracy. This approach effectively mitigates the limitations of individual models, achieving a J\&F score of 0.8637 and ranking 1st place in the MOSEv1 track of the LSVOS 2025 challenge.

%% file: sec/2_method.tex
\section{Method}

Our training framework is illustrated in Figure \ref{fig:training}. Considering the specific characteristics of the MOSE dataset, such as frequent object disappearance and reappearance, heavy occlusions, and the presence of small and visually similar objects, we fine-tune the SAM2 model on the MOSE training set to capture these complex patterns. During inference, we introduce a confidence-guided fusion segmentation strategy, which first employs a pixel-level check mechanism to identify confident foreground pixels and then applies a voting mechanism to resolve regions with inconsistent predictions for different object IDs, thereby generating reliable results across video frames, as illustrated in the framework of the inference stage in Figure \ref{fig:test}. The detailed procedure of each stage is elaborated in the following sections.
\subsection{Training}
As illustrated in Fig \ref{fig:training}, we adopt SAM2, a strong baseline model in the video object segmentation, serves as the foundation of our approach. The core mechanism of SAM2 lies in its memory attention module, which facilitates efficient cross-frame attention interactions and improves performance in object tracking and segmentation. Specifically, SAM2 first employs a MAE pretrained Hiera image encoder to extract rich frame-level feature representations. These frame embeddings are subsequently combined with historical frame features and object pointers to compute cross-attention, producing temporally consistent frame representations. The resulting features are then passed through a decoder to generate  segmentation masks. In parallel, the memory encoder further encodes and stores frame features, providing effective contextual information to guide accurate segmentation in subsequent frames.

Our training strategy is structured as follows. To maintain the generalization capability of SAM2 and mitigate the risk of overfitting, the image encoder is frozen, while the remaining components of the model are fine-tuned on the MOSE dataset. The model is initialized from the large checkpoint of SAM2 version 2.1 and fine-tuned for 40 epochs. During training, we employ a diverse set of data augmentation techniques, including RandomHorizontalFlip, RandomAffine, RandomResize, ColorJitter, and RandomGrayscale, to simulate the complex variations encountered in real-world scenarios, such as changes in motion patterns, occlusions, and lighting conditions. These augmentations enhance the model’s robustness and its ability to generalize to challenging video sequences.

To enhance segmentation performance under complex scenarios, we adopt a multi-task loss function that simultaneously supervises both pixel-level and frame-level objectives. Specifically, the pixel-level supervision comprises three loss terms: Focal Loss, which identify the foreground–background pixels and adaptively emphasizes pixels that are difficult to classify, improving the model’s ability to handle challenging regions; Dice Loss measures the region overlap between predicted masks and ground-truth annotations, which is sensitive to small object regions; and IoU Loss, which assesses the overall consistency between predicted and ground-truth masks, emphasizing holistic segmentation quality. In addition, a frame-level Classification Loss is incorporated to predict the presence or absence of target objects in each frame, providing global guidance. The combined loss function is defined as:
\begin{equation}
\mathcal{L}_{total} = \lambda_1 \mathcal{L}_{focal} + \lambda_2 \mathcal{L}_{dice} + \lambda_3 \mathcal{L}_{iou} + \lambda_4 \mathcal{L}_{cls},
\end{equation}
where $\lambda_1, \lambda_2, \lambda_3, \lambda_4$ are weighting coefficients that balance the contributions of each loss term. This multi-task optimization objectives encourages the model to capture both fine-grained pixel-level details and high-level frame-wise object presence, thereby enhancing segmentation accuracy in complex video sequences. 

\subsection{Inference}
Observing that different VOS models exhibit complementary strengths in handling various challenges, such as occlusions, small or visually similar objects, and long-term reappearances, we propose a confidence-guided multi-model ensemble strategy to leverage their individual advantages and enhance segmentation robustness. During inference, this strategy is executed in two main phases: single-model inference and multi-model fusion.

\textbf{Phase 1: Single-Model Inference.} For initial inference, we employ five models, SAM2Long, SAM2, Cutie, LiVOS, and XMem, leveraging their complementary strengths to enhance segmentation robustness. Sam2long explicitly addresses segmentation uncertainty through a Constrained Tree Search mechanism, which selects the globally optimal segmentation path across multiple candidates for the entire video. In our experiments, the parameters are set as num\_pathway to 3, iou\_thre to 0.1, and uncertainty to 1.5. For Cutie, Livos, and XMem, we apply different memory configurations to accommodate videos of varying lengths. For sequences longer than 200 frames, we use max\_mem\_frames=45, min\_mem\_frames=40, and topk=50, whereas for shorter sequences with fewer than 200 frames, we use max\_mem\_frames=15, min\_mem\_frames=14, and topk=40, thereby enhancing the model's tracking and segmentation capabilities in long-video scenarios. Additionally, to further enhance the robustness of predictions, we employ a Test-Time Augmentation (TTA) strategy, which fuses predictions from both the original and horizontally flipped frames to generate the final masks.  

\textbf{Phase 2: Multi-Model Fusion.}  
To integrate the strengths of different models across various scenarios, we generate final results by aggregating the outputs of different models. Specifically, pixel-level foreground decisions are determined by aggregating confidence scores across models, such that a pixel is classified as foreground if its cumulative score exceeds a predefined threshold. At the object level, a voting mechanism resolves prediction inconsistencies of object IDs among multiple models, producing globally consistent results. Through this approach, we can effectively address challenges arising from overlapping objects and mitigate target disappearance in long video sequences.
  
\label{sec:formatting}

%% file: sec/3_experiments.tex
\section{Experiments}
\label{sec:exp}

\begin{figure}[t]
\centering \includegraphics[width=0.48\textwidth]{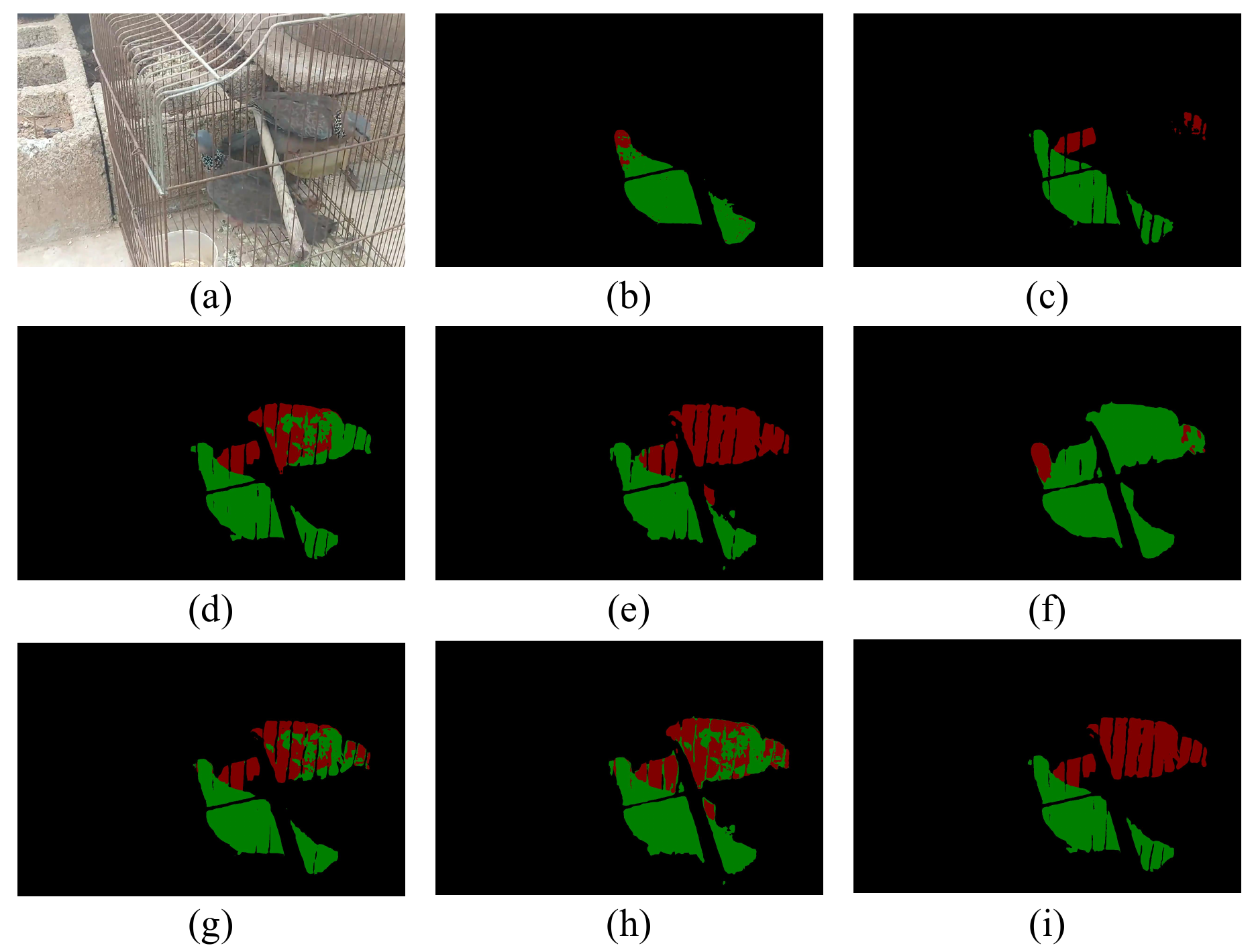} \caption{We compare the inference results of different methods on the video cMWGYYW3. (a) represents the original image. (b)--(f) show the inference results of the SAM2Long, SAM2, Cutie, LiVOS, and XMem models, respectively. (g) denotes the result obtained by averaging the masks from the Model Zoo. (h) denotes the result obtained by taking the maximum value fusion. (i) denotes the result obtained by confidence-guided fusion strategy. Here, the Model Zoo consists of five models: SAM2Long, SAM2, Cutie, LiVOS, and XMem.} \label{fig:fusion_results} 
\end{figure}

\begin{figure*}[t]
\centering \includegraphics[width=\textwidth]{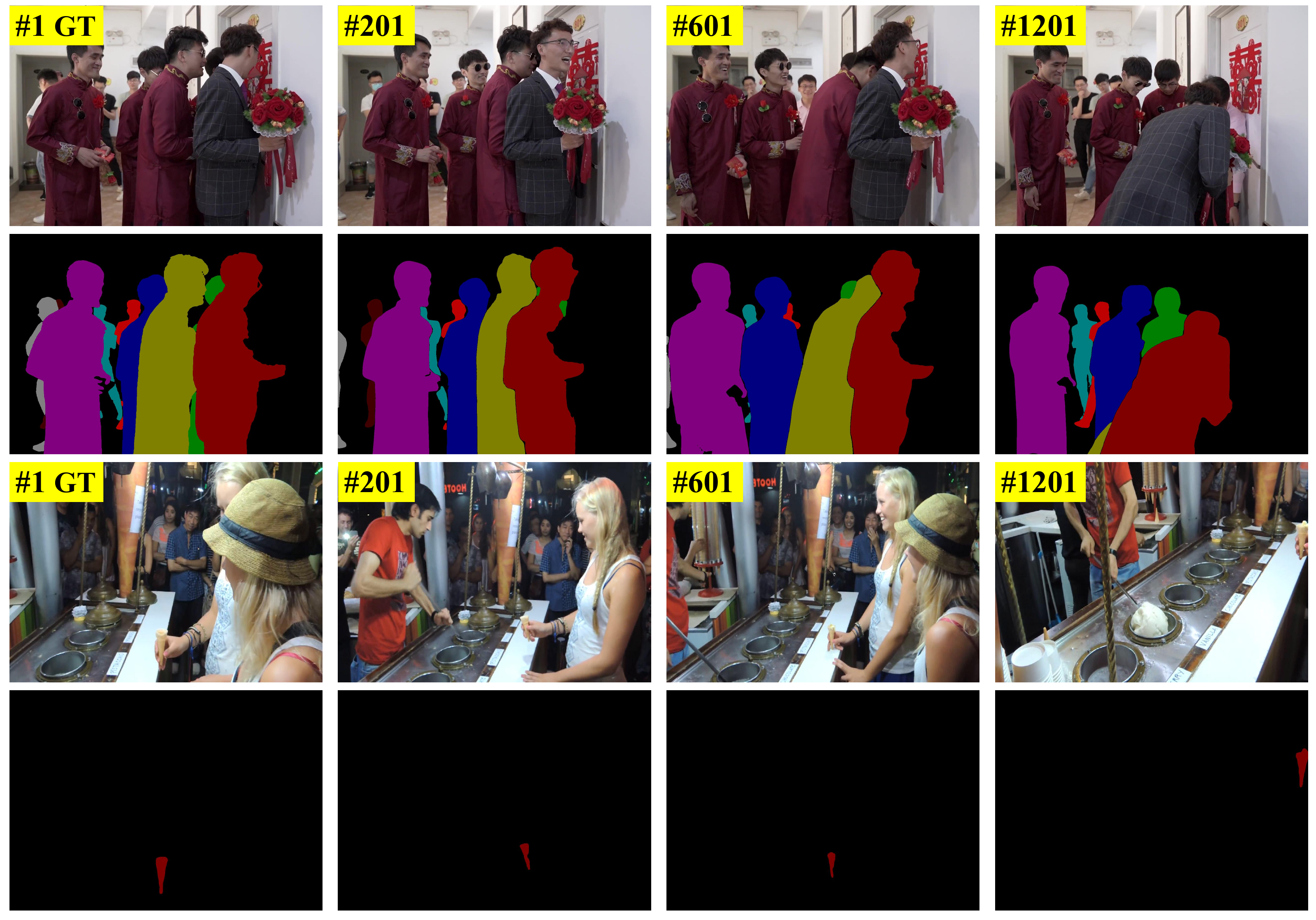} \caption{Qualitative results on complex scenes from the test set. Examples demonstrate the effectiveness of our method under challenging conditions, including heavy occlusion and small object motion.} \label{fig:results} 
\end{figure*}

\subsection{Dataset}
\label{sec:dataset}
The dataset provided for this track is a hybrid of MOSEv1 and LVOS datasets. MOSEv1 is specifically designed for VOS task in complex scenes, focusing on challenging cases such as frequent object disappearance and reappearance, small or inconspicuous objects, heavy occlusions, and crowded environments \cite{ding2025mosev2challengingdatasetvideo}. It consists of 2,149 video clips that encompass 5,200 objects from 36 categories, accompanied by 431,725 high-quality instance segmentation masks. MOSEv1 substantially exceeds existing mainstream VOS datasets in scale, as measured by the number of videos, total annotations, and duration. Furthermore, it exhibits markedly higher Mean Bounding-box Occlusion Rate (mBOR) and disappearance rate (Disapp. Rate) compared to conventional benchmarks such as DAVIS and YouTube-VOS, indicating more pervasive occlusion and more frequent object vanishing events \cite{ding2023mose}. LVOS is designed for long-term videos, aims to provide a benchmark for the development of long-term VOS models.

\subsection{Implementation Details}
\textbf{Training.} 
We employ the AdamW optimizer with a base learning rate of $5 \times 10^{-6}$ and a vision-specific learning rate of $3 \times 10^{-6}$. Input images are resized to a resolution of $768 \times 768$, and sequences of 8 frames, i.e., num\_frames=8, are used as input. The model is trained for a total of 40 epochs, requiring approximately 20 hours. All experiments are conducted on four NVIDIA A6000 GPUs.

\subsection{Evaluation Metrics}
\label{sec:Evaluation Metrics}
We compute the Jaccard value (J), the F-Measure (F), and the mean of J and F as the primary evaluation metrics. 

The Jaccard value quantifies the segmentation accuracy by measuring the overlap between the predicted and ground-truth masks, which is defined as:
\begin{equation}
J = \frac{ | {P} \cap {G} | }{ | {P} \cup {G} | },
\end{equation}
where \({P}\) denotes the predicted segmentation mask and \({G}\) represents the ground truth segmentation mask. The F-Measure provides a balanced assessment of precision and recall, emphasizing both accuracy and completeness of the details. It can be expressed as:
\begin{equation}
F = \frac{ 2 \cdot \text{Precision} \cdot \text{Recall} }{ \text{Precision} + \text{Recall} }.
\end{equation}
Precision and Recall are defined as:
\begin{align}
\text{Precision} & = \frac{ \text{TP} }{ \text{TP} + \text{FP} }, \\
\text{Recall}    & = \frac{ \text{TP} }{ \text{TP} + \text{FN} },
\end{align}
where TP, FP, and FN represent true positives, false positives, and false negatives, respectively.
Their average, J\&F, offers an overall performance measure by equally weighting both aspects, which is given below:
\begin{equation}
\text{Mean}(J,F) = \frac{J + F}{2}.
\end{equation}

\subsection{Results}
\begin{table}[t]
  \centering
  \caption{Ranking results (Top 5) in the MOSE test set.}
  \begin{tabular}{c c c c c}
    \hline
    Rank & Name & $\mathcal{J}\&\mathcal{F}$ & $\mathcal{J}$ & $\mathcal{F}$ \\
    \hline
    1 & \textcolor{black}{syyyy} & \textcolor{black}{0.8637} & \textcolor{black}{0.8410} & \textcolor{black}{0.8864} \\
    2 & isyanan1024 & 0.8616 & 0.8372 & 0.8859 \\
    3 & rookie7777 & 0.8584 & 0.8357 & 0.8810 \\
    4 & MGM & 0.8472 & 0.8239 & 0.8705 \\
    5 & hhding & 0.8447 & 0.8214 & 0.8680 \\
    \hline
  \end{tabular}
  \label{tab:results}
\end{table}

Table~\ref{tab:results} presents the leaderboard of the MOSEv1 track on the test set. Our method achieves a $\mathcal{J}$ score of 0.8410, an $\mathcal{F}$ score of 0.8864, and a combined $\mathcal{J}\&\mathcal{F}$ score of 0.8637, ranking first in the VOS track.
Additionally, Figure~\ref{fig:results} presents qualitative results on complex scenes. For instance, the first row demonstrates multiple visually similar targets in a densely crowded state with occlusions and interactions. The corresponding masks illustrate our method's multi-object segmentation performance under heavy occlusion and visual similarity challenges. The third row features frames selected from the same video spanning a longer time interval. These frames contain minute-scale targets exhibiting significant motion. Our method achieves robust and precise segmentation under the challenges of small-target motion and long-term tracking. These results validate the effectiveness of our approach in capturing complex object motions and long-term disappearances and reappearances.

\subsection{Visualization of Mask from Different Methods.}
We conduct visualization experiments to compare the masks generated by different models. As illustrated in Figure \ref{fig:fusion_results}, the confidence-guided fusion approach achieves superior segmentation performance, whereas the average-based and max-based fusion strategies yield comparatively weaker results. These findings indicate that fine-grained mask refinement guided by confidence scores can produce more accurate and reliable results.

%% file: sec/4_conclusion.tex
\section{Conclusion}
In this paper, we introduce CGFSeg, a VOS approach that integrates a confidence-guided fusion strategy with fine-tuning of the SAM2 model on the MOSE dataset. By leveraging the complementary advantages of multiple segmentation models and adapting SAM2 to capture complex motion dynamics, the proposed method effectively addresses key challenges, including small objects, long-term object reappearances, and overlapping instances. Extensive experiments validate the effectiveness of our approach, which achieved 1st place in the MOSEv1 track of the LSVOS 2025 Challenge with a J\&F score of 86.37\%, demonstrating both the robustness and accuracy of the proposed approach.

%% file: main.bbl
\begin{thebibliography}{17}
\providecommand{\natexlab}[1]{#1}
\providecommand{\url}[1]{\texttt{#1}}
\expandafter\ifx\csname urlstyle\endcsname\relax
  \providecommand{\doi}[1]{doi: #1}\else
  \providecommand{\doi}{doi: \begingroup \urlstyle{rm}\Url}\fi

\bibitem[Caelles et~al.(2017)Caelles, Maninis, Pont{-}Tuset, Leal{-}Taix{\'{e}}, Cremers, and Gool]{one_shot:conf/cvpr/CaellesMPLCG17}
Sergi Caelles, Kevis{-}Kokitsi Maninis, Jordi Pont{-}Tuset, Laura Leal{-}Taix{\'{e}}, Daniel Cremers, and Luc~Van Gool.
\newblock One-shot video object segmentation.
\newblock In \emph{CVPR}, pages 5320--5329, 2017.

\bibitem[Cheng and Schwing(2022)]{cheng2022xmem}
Ho~Kei Cheng and Alexander~G. Schwing.
\newblock Xmem: Long-term video object segmentation with an atkinson-shiffrin memory model.
\newblock In \emph{ECCV}, pages 640--658, 2022.

\bibitem[Cheng et~al.(2021)Cheng, Tai, and Tang]{memort_method_2:conf/nips/ChengTT21}
Ho~Kei Cheng, Yu{-}Wing Tai, and Chi{-}Keung Tang.
\newblock Rethinking space-time networks with improved memory coverage for efficient video object segmentation.
\newblock In \emph{NeurIPS}, pages 11781--11794, 2021.

\bibitem[Cheng et~al.(2024)Cheng, Oh, Price, Lee, and Schwing]{cheng2024putting}
Ho~Kei Cheng, Seoung~Wug Oh, Brian~L. Price, Joon{-}Young Lee, and Alexander~G. Schwing.
\newblock Putting the object back into video object segmentation.
\newblock In \emph{CVPR}, pages 3151--3161, 2024.

\bibitem[Ding et~al.(2023{\natexlab{a}})Ding, Liu, He, Jiang, and Loy]{MeViS}
Henghui Ding, Chang Liu, Shuting He, Xudong Jiang, and Chen~Change Loy.
\newblock {MeViS}: A large-scale benchmark for video segmentation with motion expressions.
\newblock In \emph{ICCV}, 2023{\natexlab{a}}.

\bibitem[Ding et~al.(2023{\natexlab{b}})Ding, Liu, He, Jiang, Torr, and Bai]{ding2023mose}
Henghui Ding, Chang Liu, Shuting He, Xudong Jiang, Philip~HS Torr, and Song Bai.
\newblock {MOSE}: A new dataset for video object segmentation in complex scenes.
\newblock In \emph{ICCV}, 2023{\natexlab{b}}.

\bibitem[Ding et~al.(2025{\natexlab{a}})Ding, Liu, He, Ying, Jiang, Loy, and Jiang]{ding2025mevis}
Henghui Ding, Chang Liu, Shuting He, Kaining Ying, Xudong Jiang, Chen~Change Loy, and Yu-Gang Jiang.
\newblock Mevis: A multi-modal dataset for referring motion expression video segmentation.
\newblock \emph{IEEE Transactions on Pattern Analysis and Machine Intelligence}, 2025{\natexlab{a}}.

\bibitem[Ding et~al.(2025{\natexlab{b}})Ding, Ying, Liu, He, Jiang, Jiang, Torr, and Bai]{ding2025mosev2challengingdatasetvideo}
Henghui Ding, Kaining Ying, Chang Liu, Shuting He, Xudong Jiang, Yu-Gang Jiang, Philip~HS Torr, and Song Bai.
\newblock {MOSEv2}: A more challenging dataset for video object segmentation in complex scenes.
\newblock \emph{arXiv preprint arXiv:2508.05630}, 2025{\natexlab{b}}.

\bibitem[Ding et~al.(2024)Ding, Qian, Dong, Zhang, Zang, Cao, Guo, Lin, and Wang]{ding2025sam2long}
Shuangrui Ding, Rui Qian, Xiaoyi Dong, Pan Zhang, Yuhang Zang, Yuhang Cao, Yuwei Guo, Dahua Lin, and Jiaqi Wang.
\newblock Sam2long: Enhancing {SAM} 2 for long video segmentation with a training-free memory tree.
\newblock \emph{CoRR}, abs/2410.16268, 2024.

\bibitem[Hong et~al.(2023)Hong, Chen, Liu, Zhang, Guo, Chen, and Zhang]{Hong_2023_ICCV}
Lingyi Hong, Wenchao Chen, Zhongying Liu, Wei Zhang, Pinxue Guo, Zhaoyu Chen, and Wenqiang Zhang.
\newblock Lvos: A benchmark for long-term video object segmentation.
\newblock In \emph{Proceedings of the IEEE/CVF International Conference on Computer Vision (ICCV)}, pages 13480--13492, 2023.

\bibitem[Oh et~al.(2019)Oh, Lee, Xu, and Kim]{oh2019stm}
Seoung~Wug Oh, Joon{-}Young Lee, Ning Xu, and Seon~Joo Kim.
\newblock Video object segmentation using space-time memory networks.
\newblock In \emph{ICCV}, pages 9225--9234, 2019.

\bibitem[Perazzi et~al.(2016)Perazzi, Pont{-}Tuset, McWilliams, Gool, Gross, and Sorkine{-}Hornung]{Benchmark:conf/cvpr/PerazziPMGGS16}
Federico Perazzi, Jordi Pont{-}Tuset, Brian McWilliams, Luc~Van Gool, Markus~H. Gross, and Alexander Sorkine{-}Hornung.
\newblock A benchmark dataset and evaluation methodology for video object segmentation.
\newblock In \emph{CVPR}, pages 724--732, 2016.

\bibitem[Pont{-}Tuset et~al.(2017)Pont{-}Tuset, Perazzi, Caelles, Arbel{\'{a}}ez, Sorkine{-}Hornung, and Gool]{davis17:journals/corr/Pont-TusetPCASG17}
Jordi Pont{-}Tuset, Federico Perazzi, Sergi Caelles, Pablo Arbel{\'{a}}ez, Alexander Sorkine{-}Hornung, and Luc~Van Gool.
\newblock The 2017 {DAVIS} challenge on video object segmentation.
\newblock \emph{CoRR}, abs/1704.00675, 2017.

\bibitem[Ravi et~al.(2025)Ravi, Gabeur, Hu, Hu, Ryali, Ma, Khedr, R{\"{a}}dle, Rolland, Gustafson, Mintun, Pan, Alwala, Carion, Wu, Girshick, Doll{\'{a}}r, and Feichtenhofer]{ravi2024sam2}
Nikhila Ravi, Valentin Gabeur, Yuan{-}Ting Hu, Ronghang Hu, Chaitanya Ryali, Tengyu Ma, Haitham Khedr, Roman R{\"{a}}dle, Chlo{\'{e}} Rolland, Laura Gustafson, Eric Mintun, Junting Pan, Kalyan~Vasudev Alwala, Nicolas Carion, Chao{-}Yuan Wu, Ross~B. Girshick, Piotr Doll{\'{a}}r, and Christoph Feichtenhofer.
\newblock {SAM} 2: Segment anything in images and videos.
\newblock In \emph{ICLR}, 2025.

\bibitem[Seong et~al.(2020)Seong, Hyun, and Kim]{memort_method_3:conf/eccv/SeongHK20}
Hongje Seong, Junhyuk Hyun, and Euntai Kim.
\newblock Kernelized memory network for video object segmentation.
\newblock In \emph{ECCV}, pages 629--645, 2020.

\bibitem[Videnovic et~al.(2025)Videnovic, Lukezic, and Kristan]{distractor:conf/cvpr/VidenovicLK25}
Jovana Videnovic, Alan Lukezic, and Matej Kristan.
\newblock A distractor-aware memory for visual object tracking with {SAM2}.
\newblock In \emph{CVPR}, pages 24255--24264, 2025.

\bibitem[Xu et~al.(2018)Xu, Yang, Fan, Yue, Liang, Yang, and Huang]{xu2018youtube}
Ning Xu, Linjie Yang, Yuchen Fan, Dingcheng Yue, Yuchen Liang, Jianchao Yang, and Thomas~S. Huang.
\newblock Youtube-vos: {A} large-scale video object segmentation benchmark.
\newblock \emph{CoRR}, abs/1809.03327, 2018.

\end{thebibliography}
